\documentclass[letterpaper, 10 pt, conference]{ieeeconf}
\usepackage{graphicx} % Required for the inclusion of images
\usepackage{hyperref}
\usepackage{subcaption}
\usepackage{colonequals}
\usepackage{amsmath} % Required for some math elements
\usepackage{float}
\usepackage{booktabs}
\usepackage{amsfonts}
\usepackage{amssymb}
\usepackage[dvipsnames]{xcolor}
\usepackage{siunitx}
\usepackage{bm}
\usepackage[font=small,labelfont=bf]{caption}

\usepackage{MnSymbol}

\usepackage[ruled]{algorithm2e}
\usepackage{algorithmic}
\usepackage{multirow}
\usepackage{tabularx}
\usepackage{lipsum}
\usepackage{float}
\makeatletter
\newcommand{\algorithmstyle}[1]{\renewcommand{\algocf@style}{#1}}
\newcommand{\removelatexerror}{\let\@latex@error\@gobble}

\makeatother
\usepackage[noadjust]{cite}

\newcommand{\ours}[0]{\textit{RA-DP}} 
\IEEEoverridecommandlockouts                              % This command is only needed if 
\title{\LARGE \bf
\ours: \underline{R}apid \underline{A}daptive \underline{D}iffusion \underline{P}olicy for Training-Free High-frequency Robotics Replanning}

\author{Xi Ye$^{* \ddagger 1}$, Rui Heng Yang$^{* \dagger 2}$, Jun Jin$^3$, Yinchuan Li$^{4}$, Amir Rasouli$^{2}$% <-this % stops a space
\thanks{$^{*}$ Equal Contribution.
        $^{\dagger}$ Correspondence to: Rui Heng Yang (rui.heng.yang2@huawei.com).
        $^{1}$ Department of Computer Science and Technology, Tsinghua University.
        $^{2}$ Noah's Ark Laboratory, Huawei Technologies Canada.
        $^{3}$ Department of Electrical and Computer Engineering, University of Alberta.
        $^{4}$ Noah's Ark Laboratory, Huawei Technologies.
        $^{\ddagger}$ Work was done while at Noah's Ark Laboratory.
        }
    }
    
\begin{document}
\maketitle
\thispagestyle{empty}
\pagestyle{empty}

\begin{abstract}

% Diffusion models exhibit impressive scalability in robotic task learning, yet they struggle to adapt to novel, highly dynamic environments. This limitation primarily stems from their constrained replanning ability: they either operate at a low 
% frequency due to a time-consuming iterative sampling process, or are unable to adapt to unforeseen feedback in case of rapid replanning. To address these challenges, we propose \ours, a novel diffusion policy framework with training-free high-frequency replanning ability that solves the above limitations in adapting to unforeseen dynamic environments. In this work, we focus on manipulation with collision avoidance in environments where not all obstacles are foreseeable. We regard this problem as one of the most challenging tasks encountered in dynamic settings. Integrated with guidance signals from novel scenarios, in a training-free manner, \ours\space leverages an innovative action queue mechanism to generate replanned actions at every denoising step in the diffusion sampling process. Experiments across simulated and real-world goal-reaching with obstacle avoidance demonstrate that \ours\space outperforms the state-of-the-art diffusion-based methods in terms of replanning frequency and success rate. Moreover, we show that our framework is theoretically compatible with any training-free guidance signal. 

Diffusion models exhibit impressive scalability in robotic task learning, yet they struggle to adapt to novel, highly dynamic environments. This limitation primarily stems from their constrained replanning ability: they either operate at a low frequency due to a time-consuming iterative sampling process, or are unable to adapt to unforeseen feedback in case of rapid replanning. To address these challenges, we propose \ours, a novel diffusion policy framework with training-free high-frequency replanning ability that solves the above limitations by adapting to unforeseen dynamic environments. Specifically, our method integrates guidance signals, which are often easily obtained in the new environment during the diffusion sampling process, and utilizes a novel action queue mechanism to generate replanned actions at every denoising step without retraining, thus forming a complete training-free framework for robot motion adaptation in unseen environments. We conduct extensive evaluations in both common simulation benchmarks and real-world environments. Our results indicate that \ours\space outperforms the state-of-the-art diffusion-based methods in terms of replanning frequency and success rate. At the end, we show that our framework is theoretically compatible with any training-free guidance signal, hence increasing its applicability to a wide range of robotics tasks. 

\end{abstract}
%%%%%%%%%%%%%%%%%%%%%%%%%%%%%%%%%%%%%%%%%%%%%%%%%%%%%%%%%%%%%%%%%%%%%%%%%%%%%%%%
    
\section{Introduction}
\label{sec:intro}
Diffusion models have gained widespread attention for their effective performance in various tasks, such as image generation \cite{Epstein_Diff_2023_NIPS}, video synthesis \cite{Esser_2023_ICCV}, and text-to-image conversion \cite{Li_textdif_2023_NIPS}. The ability of diffusion models to learn complex data distributions has made them a powerful tool in creative and predictive applications. More recently, diffusion models have been employed for learning robotic control tasks \cite{chi2023,reuss2023goal,Ze2024DP3, prasad2024consistency}. In this context, they have demonstrated the expressiveness needed for complex manipulation tasks, stable training, and adaptation to higher-dimensional action spaces. These characteristics make diffusion models ideal for addressing a wide range of challenges in robotic control.

Diffusion-based robot controllers, however, have a limited replanning frequency, which hinders their ability to adapt to rapidly changing environments. Here, replanning frequency refers to the rate at which the controller generates a new action in response to changing environmental conditions. Such a characteristic is particularly crucial in goal reaching tasks involving fast-moving and unforeseen obstacles that can potentially cause collisions and consequently degrade performance.

\begin{figure}[t]
\vspace{0.2cm}
\centering
\includegraphics[clip, trim=0.1cm 1cm 0cm 1cm, width=\linewidth]{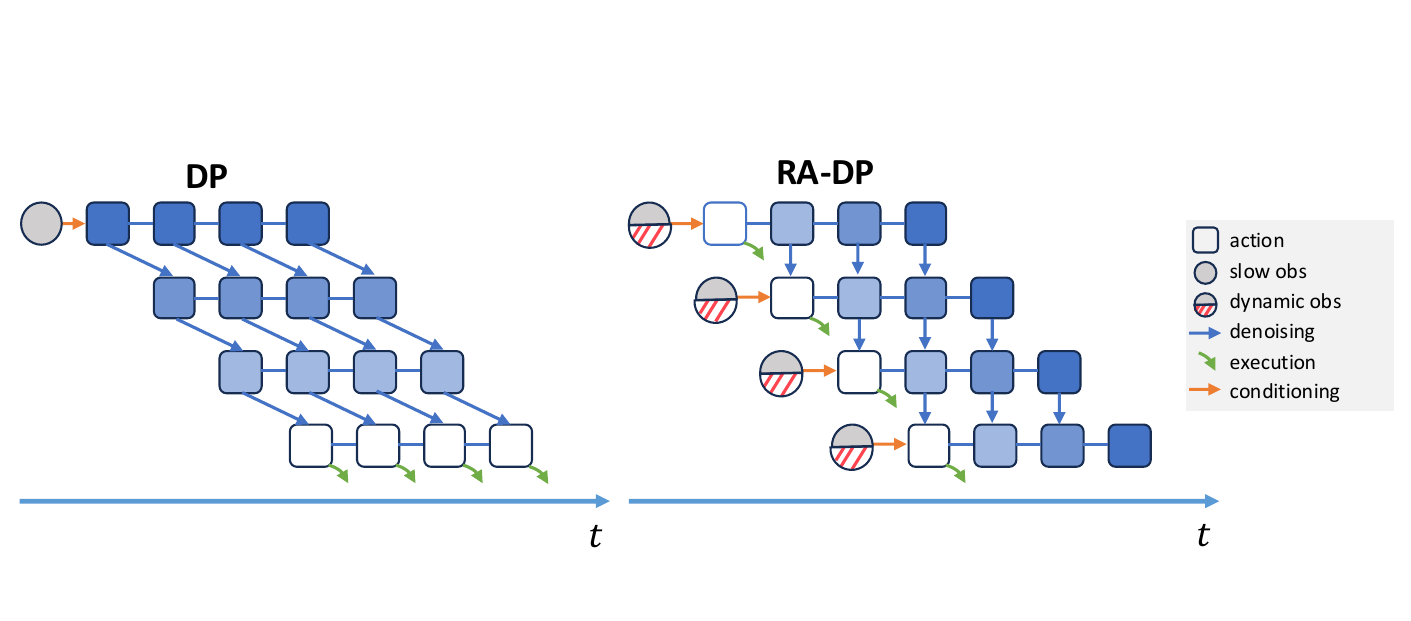}\vspace{-0.3cm}
\caption{Comparison between Diffusion Policy (DP) and \ours. DP has limited replanning frequency and fails to adapt to unforeseen dynamic observations, whereas \ours\ replans actions using the latest dynamic feedback at every  denoising step in a training-free manner. The lighter the shade of blue, the lower the level of Gaussian noise. A white action represents an executable action.}
\label{fig:teasing_fig}
\vspace{-0.5cm}
\end{figure}
For robot control, the authors of \cite{chi2023} propose Diffusion Policy (DP), which typically generates 16 actions by iterative sampling, executes the first 8 actions, and then replans conditioned on the new observations. DP has replanning frequency of $\approx 37$ Hz, which is substantially slower compared to standard low-level robotic control systems. As a result, DP can create a bottleneck for the entire system. One way to increase the replanning rate is through shortening of prediction and action horizons at the cost of lowering the success rate. Alternatively, one can resort to fast diffusion sampling techniques, such as consistency models \cite{prasad2024consistency}, to improve frequency. However, these approaches still struggle to adapt to novel environment feedback, such as unforeseen obstacles.

To improve adaptability, existing diffusion-based models incorporate environmental feedback as conditional inputs for action planning, often via classifier-free guidance \cite{reuss2023goal}. These methods, however, fail to generalize to unforeseen conditions during inference. This problem can be resolved using classifier-guided approaches \cite{janner2022diffuser}, which incorporate novel feedback by training an additional discriminator for each new feedback modality. However, besides adding computational burden for training, encoding certain feedback types as conditional variables can be deemed difficult. In the collision avoidance context, for instance, it is infeasible to include all possible obstacle interactions during training, thus the classifier-guided approaches would fail to scale.

To address these limitations, we propose \ours, an adaptive diffusion-based framework that effectively respond to unforeseen, rapidly changing environment feedback via high-frequency replanning (Fig. \ref{fig:teasing_fig}). Our approach benefits from integrating training-free loss-based guidance methods \cite{shen2024understanding, Yu_2023_ICCV}, enabling the policy to generalize to any conditional inputs-encoded as a scalar differential loss function during inference without the need for additional training. Furthermore, inspired by a recent human motion synthesis work \cite{zhang2024tedi}, \ours\space uses a novel action queue mechanism for both training and inference processes, which enables high-frequency control in dynamic environments. 

\ours\space establishes a fixed-length action queue. During training, each action within the queue is independently perturbed with varying noise levels, unlike previous methods that use uniform noise levels across actions. A standard U-Net is trained to predict clean actions conditioned on slowly changing feedback (e.g., a fixed goal position). During inference, \ours\space maintains an action queue with monotonically increasing noise levels. Other than conditioning on the foreseen, slow-changing feedback during the training stage, each action is also seamlessly steered by the training-free guidance from possible rapid changing environment feedback. \ours\space removes the executable clean action from the front of the queue (\textit{dequeue}) and appends a new noisy one to the tail at every denoising step (\textit{enqueue}), thereby achieving high-frequency training-free replanning. For example, in goal-reaching tasks in cluttered environments, \ours\space is able to incorporate live obstacle data to dynamically guide the actions.

In summary, the contributions of our work are as follows:
\begin{itemize}
    \item We propose \ours, a novel diffusion-based framework for robotic manipulation with a training-free high-frequency replanning to adapt to fast-changing environments.
    \item We conduct empirical evaluations and show that \ours\space  achieves state-of-the-art performance while maintaining high replanning frequency across diverse simulated benchmark tasks. 
    \item We demonstrate the practical application of our training-free high-frequency replanning approach in real-world goal-reaching manipulation tasks involving collision avoidance.

\end{itemize}

\section{Related works}
\label{sec:related_works}

\subsection{Diffusion Models in Robotics}
Diffusion models have been widely used in robotics \cite{pearce2023imitating, chi2023, Ze2024DP3, reuss2023goal, janner2022diffuser}, particularly for imitation learning to address the challenges of multimodal behavior and high-dimensional action spaces. Diffusion behavior cloning \cite{pearce2023imitating} and Diffusion Policy (DP) \cite{chi2023} are among the first to apply diffusion models in robotics. In this context, imitation learning is framed as a conditional generative task where actions are generated conditioned on observations. In \cite{Ze2024DP3}, DP is further extended to 3D visual representations.

While these diffusion-based approaches excel in static or slow-changing environments, they fall short in rapidly evolving real-world scenarios. This is primarily due to their limited replanning frequency, hindering their application to tasks, such as collision avoidance in dynamic settings. To address this issue, we propose a novel high-frequency replanning diffusion-based framework that benefits from a  training-free loss-based guidance in order to enhance adaptability and efficiency of the controller in highly dynamic environments.

\subsection{Guidance Mechanisms in Diffusion Models}
Vanilla diffusion models struggle to guide the action generation towards desired outcomes that maximize rewards or satisfy the constraints while maintaining high sample quality. To address this limitation, the behavior synthesis policy \cite{janner2022diffuser} employs classifier-guided diffusion \cite{dhariwal2021diffusion}, which uses a pretrained classifier to adjust the denoising process using its gradients. However, this approach requires a dedicated classifier for each conditioning modality, making it impractical for dynamically changing environments, where novel modalities may be introduced. Alternatively, some goal-conditioned diffusion policies \cite{reuss2023goal} adopt classifier-free guidance methods \cite{ho2022classifierfree} by integrating the guidance signal directly into the diffusion model during training. Despite eliminating the need for an external classifier, this approach fails to generalize to unforeseen conditioning information, such as novel object compositions, during inference.

More recent works rely on training-free, loss-based guidance methods to enable diffusion models to adapt to arbitrary conditioning signals without retraining \cite{shen2024understanding,yu2023unified}. Our work builds upon these approaches by integrating loss-based, training-free guidance into our diffusion-based framework for robot control. By enabling training-free guidance during the sampling phase, our method significantly enhances the adaptability and efficiency of the policy for complex real-world tasks.

\subsection{Fast Replanning in Diffusion-based Controllers}

Diffusion models \cite{ho2020denoising,song2021scorebased} generate high-quality samples through iterative refinement from Gaussian noise. This iterative process, however, leads to low replanning frequencies, limiting diffusion models real-time performance. To overcome this challenge, the approaches, such as DDIM \cite{song2020denoising}, DPM-Solver \cite{lu2022dpm}, and Genie \cite{dockhorn2022genie} use advanced ordinary differential equation (ODE) solvers to reduce the number of sampling steps needed for high-quality samples. Despite such an improvement, these methods  fail to achieve the high-frequency replanning required for low-latency decision-making applications.

A more recent consistency policy \cite{prasad2024consistency} leverages diffusion model distillation to enable single-step sampling via direct noise-to-data mappings. Although this approach facilitates fast replanning, it has a number of shortcomings. The additional distillation stage applied to pretrained diffusion policies increases computational cost; The single-step sampling process compromises accuracy compared to iterative approaches; Lastly, the single-step architecture lacks intermediate latent representations, precluding the application of training-free, loss-based guidance techniques \cite{Yu_2023_ICCV} necessary for adapting to unforeseen conditions. This is due to the fact that applying guidance directly to the data space either causes divergence or fails to influence the sampling process effectively.

In this work, we propose a significantly different approach, in which we integrate the action queue with different noise levels for high-frequency replanning. More specifically, our training and inference pipeline denoises the action queue with monotonically increasing noise levels, enabling single-step action generation without compromising sample quality. Additionally, we retain intermediate latent representations for training-free, loss-based guidance, thereby supporting adaptation to novel scenarios during inference. These characteristics make our proposed method compatible with any existing fast diffusion solvers. 

Concurrent works, such as streaming diffusion policy (SDP) \cite{høeg2024streamingdiffusionpolicyfast} and diffusion forcing, have implemented action queue mechanisms analogous to our method. However, SDP is open-loop control and it lacks support for training-free guidance, a capability central to our framework. Alternatively, diffusion forcing employs classifier guidance dependent on a pretrained classifier for fixed conditional variables, thus it is not training-free and cannot adapt to different feedback during inference.

Similarly, hierarchical policies, such as Yell At Your Robot (YAY) \cite{shi2024yell}, utilize high-level human language interventions as real-time corrective feedback for low-level policy errors. However, YAY's VAE-based language-conditioned behavior cloning policy exhibits constrained action generation capabilities compared to diffusion-based approaches. Furthermore, while YAY requires continuous training to integrate language feedback, our method employs a training-free guidance mechanism.

\begin{figure*}[t]
\vspace{0.2cm}
\centering
\includegraphics[width=0.9\textwidth]{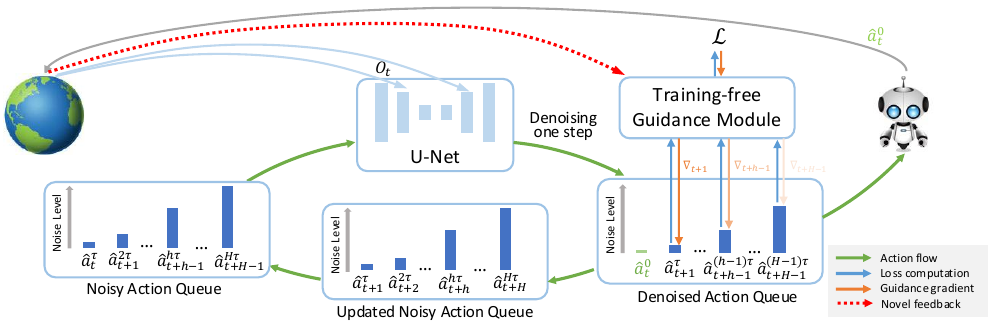}\vspace{-0.3cm}
\caption{\ours\space inference process. Starting with a noisy action queue, \ours\space performs single-step denoising conditioned on the current foreseen slow-changing observations $O_t$, producing a denoised action queue. The guidance loss ($\mathcal{L}$) is computed given the denoised action queue and dynamic novel feedback. Then training-free guidance gradients ($\nabla$) are applied, after which the first clean action $\hat{a}_t^0$ is executed and removed from the action queue (\textit{Dequeue}), while a new noisy action is appended (\textit{Enqueue}), forming an updated queue. Note that darker guidance gradients ($\nabla$) indicate stronger influence.}
\label{fig:ugcdp_model}
\vspace{-0.5cm}
\end{figure*}

\section{Proposed Method}
\label{sec:method}
We begin by introducing the notation used throughout this paper and review the key aspects of Diffusion Policy (DP) in Section \ref{ssec:dp_review}. 
Then, in Section \ref{ssec:training}, we present the training method of our \ours\space framework, which manages actions with independent noise levels during forward diffusion processes as opposed to how DP operates. Lastly, in Section \ref{ssec:inference}, we provide a detailed explanation of the inference process of our framework and demonstrate how it effectively achieves adaptive high-frequency replanning.

\subsection{Notation and Diffusion Policy (DP)}
\label{ssec:dp_review}

Given a set of expert demonstrations, let $A_t = [a_t, \dots, a_{t+H-1}]$ denote a sequence of actions over a horizon $H$ with each action $a_t \in \mathbb{R}^{d_a}$ (where $d_a$ is the action dimensionality), and let $O_t = [o_{t-N+1}, \dots, o_t]$ denote the previous $N$ observation steps with each $o_t \in \mathbb{R}^{d_o}$ (where $d_o$ is the observation dimensionality). We can employ a Denoising Diffusion Probabilistic Model (DDPM) \cite{ho2020denoising} to learn the conditional distribution $p_\theta(A_t|O_t)$, where $\theta$ represents the parameters of the diffusion network. The observations $O_t$, along with other optional conditional information are fed into the diffusion U-Net \cite{ronneberger2015u} via Feature-wise Linear Modulation (FiLM) \cite{perez2018film}.

For standard DDPM training in DP, the forward diffusion process is described as a Markov chain. At each step $k$, we have $q(A^k|A^{k-1}) = \mathcal{N}(A^k;\sqrt{1-\beta^k}A^{k-1}, \beta^k\mathbf{I})$. The scalar $\beta^k$ represents the variance of Gaussian noise at step $k$, following a prescribed schedule $\{\beta^k \in (0, 1)\}_{k=1}^K$, where $K$ denotes the maximum number of diffusion steps. Accordingly, the learned conditional reverse diffusion process at step $k$ is $p_\theta(A^{k-1}|A^k) = \mathcal{N}(A^{k-1}; \mu_\theta(A^k, k, O), \sigma^k\mathbf{I})$, where $\mu_\theta$ denotes the learned mean and $\sigma^k$ is the variance, which depends on $k$. We can reparameterize $\mu_\theta$ with a noise prediction network, $\mu_\theta(A^k, O, k) = \frac{1}{\sqrt{1-\beta^k}}(A^k - \frac{\beta^k}{\sqrt{1-\bar{\alpha}^k}}\epsilon_\theta(A^k, k, O))$, where $\theta$ denotes the neural network parameters and $\bar{\alpha}^k = \prod_{i=1}^k(1-\beta^i)$. Such reparameterization leads to a simplified MSE loss for the DDPM training in DP,
\begin{equation}
\label{loss-1}
    L=\mathbb{E}_{A^0, k, O, \epsilon \sim \mathcal{N}(0, \mathbf{I})} \lVert \epsilon - \epsilon_\theta(A^k, k, O)) \rVert ^2_2,
\end{equation}
where $A^k = \sqrt{\bar{\alpha}^k} A^0 + \sqrt{1-\bar{\alpha}^k}\epsilon$ and $\epsilon$ denotes a random Gaussian noise.

We emphasize that the standard DDPM training used in diffusion treats the entire prediction sequence $A_t$ as a single example, where each action $a_t$ in the sequence is perturbed equally with the same noise level during the forward diffusion process. Instead, if we sample a sequence of actions during inference with varying noise levels, the policy trained with standard DDPM diverges and fails to predict normal actions. This is due to the mismatch between training and inference paradigms. However, we argue that varying noise level is needed to achieve real-time close-loop control.

\subsection{The Training Process}
\label{ssec:training}

To achieve high-frequency replanning with our \ours\space framework, we adopt a new training paradigm. Instead of perturbing all actions in sequence $A_t$ by noise with the same variance $\beta^k$, we independently sample $H$ different variance values from $\{\beta^k \in (0, 1)\}_{k=1}^K$, the corresponding diffusion steps are denoted as $\mathbf{k} \in \mathbb{R}^H$. In this way, the MSE loss in Eq. \eqref{loss-1} becomes,
\begin{equation}
\label{loss-2}
L=\mathbb{E}_{A^0, \mathbf{k}, O, \epsilon \sim \mathcal{N}(0, \mathbf{I})} \lVert \epsilon - \epsilon_\theta(A^\mathbf{k}, \mathbf{k}, O)) \rVert ^2_2,
\end{equation}
where $A^\mathbf{k} = \sqrt{\bar{\alpha}^\mathbf{k}} A^0 + \sqrt{1-\bar{\alpha}^\mathbf{k}}\epsilon$.

In practice, however, we found that employing the equivalent training loss to directly predict clean action $A^0$ instead of noise $\epsilon$ achieves a better convergence and performance during inference. In this case, the diffusion U-Net serves as a sample predictor $x_\theta$, hence, the final training loss used in our framework can be defined as,
\begin{equation}
L=\mathbb{E}_{A^0, \mathbf{k}, O} \lVert A^0 - x_\theta(A^\mathbf{k}, \mathbf{k}, O)) \rVert ^2_2.
\label{eq:train_loss}
\end{equation}

To incorporate the varying noise levels $\mathbf{k}$, we modify the vanilla DP's Conv1D U-Net architecture by adding a downsampling layer/upsampling layer for diffusion step embedding in each downsampling/upsampling U-Net block. During training, we randomly perturb actions using either independent or special monotonically increasing noise levels that align with our inference process (described in Section \ref{ssec:inference}), consequently leading to better performance. The impact of mixed noise schedules is investigated in the ablation study (Section \ref{ssec:ablation}).

\subsection{Adaptive High-frequency Replanning Process}
\label{ssec:inference}

With the proposed \ours\space  model trained using the aforementioned noise schedule and the loss function in Eq. \eqref{eq:train_loss}, we perform reverse diffusion on noisy actions with various noise levels, thereby enabling our specialized action queue sampling. Fig. \ref{fig:ugcdp_model} illustrates the complete sampling process along with the corresponding action queue updates. We now delve into the details.

\textbf{Overview.} Given the initial observation $O_0$ and pure noisy initial action $\hat{A}_0^K \sim \mathcal{N}(0, \mathbf{I})$, we perform a normal diffusion sampling using DDIM with $H$ sampling steps (\textbf{equal to the prediction horizon}). Then, we store all the intermediate latent actions $[\hat{A}_0^{\tau}, \hat{A}_0^{2\tau}, ..., \hat{A}_0^{H\tau=K}]$, where each $\hat{A}_0 \in R^{H\times d_a}$, and $\tau=\frac{K}{H}$ denotes the gap between two adjacent reverse diffusion steps.

Next, we initialize an action queue $\hat{A}_0^{\mathbf{k}} = [\hat{a}_0^{\tau}, \hat{a}_1^{2\tau}, ..., \hat{a}_{H-1}^{H\tau}]$ by taking one action $\hat{a}_{h-1}^{h\tau} \in R^{d_a}$ from each stored intermediate latent action. Here, $\mathbf{k}=[\tau, 2\tau, ..., H\tau]$ is defined as a monotonically increasing noise level. For any action queue $\hat{A}_t^{\mathbf{k}}$ at time step $t$, we conduct a one step diffusion sampling process using DDIM, resulting in $\hat{A}_t^{\mathbf{k} - 1} = [\hat{a}_t^0, \hat{a}_{t+1}^{\tau}, ..., \hat{a}_{t+H-1}^{(H-1)\tau}]$, where the noiseless first action $\hat{a}_t^0$ will be taken out from the queue for execution. 

Finally, we update $\hat{A}_t^{\mathbf{k-1}}$ by appending a new noisy action $\hat{a}_{t+H}^{H\tau}$ to get the new action queue $\hat{A}_{t+1}^\mathbf{k}$. Leveraging our proposed action queue sampling method, at each time step $t$ after just one denoising step, we obtain a noiseless executable action $\hat{a}_t^0$ conditioned on the most recent observation $O_t$. 

\textbf{High-Frequency Replanning.} Let $\Delta_p$ denote the cost of a one-step DDIM sampling and $\Delta_a$ the cost of a one-step action execution. Then, the replanning frequency of our framework is given by $\frac{1}{\Delta_p + \Delta_a}$. Notably, our method preserves sampling quality unlike other single-step diffusion samplers, as the total number of sampling steps for each action remains $H$.

In contrast, a vanilla DP with the same number of DDIM sampling steps $H$ firstly takes $H\Delta_p$ for prediction, and then $H_a\Delta_a$ for execution before receiving the next new observation, where $H_a$ is the action horizon. Consequently, its replanning frequency is given by $\frac{1}{H\Delta_p + H_a\Delta_a}$, which is considerably slower compared to our method. More importantly, our action queue sampling method is compatible with any diffusion sampler and it is not limited only to DDIM.

\textbf{Adaptation to Novel Environment Feedback.} Beyond conditioning on the slow-changing observation $O_t$,  high-frequency replanning frameworks must swiftly respond to unforeseen environmental feedback. To this end, we propose to integrate a training-free loss-based guidance module \cite{yu2023freedom, bansal2023universal, shen2024understanding}, which outputs guidance gradients from arbitrary predefined constraints that influences the diffusion sampling process, thus achieving adaptation to novel environment feedback (see Fig.\ref{fig:ugcdp_model}). Moreover, the guidance module can be seamlessly integrated with the aforementioned action queue mechanism. 

We describe the guided diffusion model from the perspective of score matching-based generative model. Here, both the forward and reverse diffusion processes are formulated as stochastic differential equations (SDEs), for which we train the U-Net to parameterize the score function $\nabla_{A^\mathbf{k}}\log p_\theta(A^\mathbf{k})$. However, a guided sampling process aims to sample $A^\mathbf{k}$ conditioned on the guidance signal $\mathcal{G}$, i.e., it requires the following score function:
\begin{equation}
    \nabla_{A^\mathbf{k}}\log p_\theta(A^\mathbf{k}|\mathcal{G}) = \nabla_{A^\mathbf{k}}\log p_\theta(A^\mathbf{k}) + \nabla_{A^\mathbf{k}}\log p_\theta(\mathcal{G}|A^\mathbf{k}),
\label{eq:guided_score_function}
\end{equation}
where the first term in the RHS can be provided by the trained DP model. For the second term in Eq. \eqref{eq:guided_score_function}, both classifier-free guidance and classifier-guidance require pairing the dataset with conditional variable $\mathcal{G}$ during training. Then it can be derived either by learning a $\mathbf{k}$-dependent classifier or randomly feeding $\mathcal{G}$ into the diffusion U-Net as an additional conditional variable. Neither of these methods, however, are training-free or are able to adapt to unforeseen $\mathcal{G}$.

Instead, we employ an energy function-based distribution $p_\theta(A^\mathbf{k}|\mathcal{G}) = p_\theta(A^\mathbf{k})e^{-f(A^\mathbf{k}, \mathcal{G})}$ 
to describe the desired preference conditioned on action distribution, where $f$ is the energy function that outputs the guidance loss given $\mathcal{G}$ and $A^\mathbf{k}$. This distribution can be interpreted as the intersection of the high density region of both $p_\theta(A^\mathbf{k})$ and preference $e^{-f(A^\mathbf{k}, \mathcal{G})}$ \cite{lu2023contrastive}. In this case, the second term in RHS of Eq.,
\eqref{eq:guided_score_function} can be derived as,
\begin{equation}
\nabla_{A^\mathbf{k}}\log p_\theta(\mathcal{G}|A^\mathbf{k}) = \nabla_{A^\mathbf{k}} \log \mathbb{E}_{p_\theta(\hat{A}^0|A^\mathbf{k})}[e^{-f(\hat{A}^0, \mathcal{G})}],
\label{eq:energy_based_guidance_grad}
\end{equation}
which is intractable to compute and requires an approximation method  \cite{lu2023contrastive, bansal2023universal, shen2024understanding, Yu_2023_ICCV, song2023loss}. Here, we approximate the training free guidance gradient following \cite{shen2024understanding, song2023loss},
\begin{equation}
\nabla_{A^\mathbf{k}}\log p_\theta(\mathcal{G}|A^\mathbf{k}) \approx -\nabla_{A^\mathbf{k}} f(\hat{A}^0, \mathcal{G}).
\label{eq:approx_guidance_grad}
\end{equation}

Given the predicted $\hat{A}^0$ and guidance loss $\mathcal{L}=f(\hat{A}^0, \mathcal{G})$, we backpropagate through loss function $f$ and the diffusion U-Net to compute the guidance gradient, which is then applied to adjust the sampled $A^{\mathbf{k} - 1}$. To enhance guided sampling convergence, the guidance gradient is weighted using the Polyak step size \cite{shen2024understanding}. For detailed guidance application, see Line 5 in Algorithm \ref{alg:inference_algorithm}. 

Although the same guidance step size is used at different time steps, the previous work \cite{yu2023freedom} has shown that guidance has limited influence on intermediate samples with higher noise levels. Consequently, in our action queue with a monotonically increasing noise level, future actions receive reduced effective guidance from the current time step, which is desirable for replanning. The entire adaptive high-frequency replanning with training-free guidance is shown in Fig.\ref{fig:ugcdp_model} and inference process is detailed in Algorithm \ref{alg:inference_algorithm}.

\begin{algorithm}
\caption{Adaptive High-frequency Replanning}\label{alg:inference_algorithm}
\hspace*{\algorithmicindent} \textbf{Input} Observations: $O_0$ \\
\hspace*{\algorithmicindent} \textbf{Initialize} $\mathbf{k}=[\tau, 2\tau, ..., H\tau], \tau=\frac{K}{H}$, step size $\eta$ \\
\hspace*{\algorithmicindent} \textbf{Initialize} Action queue: $A_0^{\mathbf{k}} = [\hat{a}_0^{\tau}, \hat{a}_1^{2\tau}, ..., \hat{a}_{H-1}^{H\tau}]$ \\ 

\hspace*{\algorithmicindent} \textbf{Initialize} Diffusion network $x_\theta$ \\

\hspace*{\algorithmicindent} \textbf{Define} Differentiable guidance loss function $f$

\begin{algorithmic}[1]

\FOR{$t = 0, 1, ..., T$}

\STATE Inference diffusion U-Net: $\hat{A}_t^0 = x_\theta(A_t^\mathbf{k}, \mathbf{k}, O_t)$
  
\STATE $A_t^{\mathbf{k}-1} = \text{DDIM}(\hat{A}_t^0)$

\STATE $g_t = \nabla_{A^{\mathbf{k}}_t}f(\hat{A}_t^0, \mathcal{G})$ according to Eq. \eqref{eq:approx_guidance_grad}

\STATE $A_t^{\mathbf{k}-1} = A_t^{\mathbf{k}-1} - \eta\cdot \frac{g_t}{\lVert g_t \rVert_2^2}$

\STATE Execute action $\hat{a}^0_t$, receive new observation $O_{t+1}$

\STATE Update action queue: $A_{t+1}^{\mathbf{k}} = [\hat{a}_{t+1}^{\tau}, ..., \hat{a}_{t+H}^{H\tau}]$

\ENDFOR
\end{algorithmic}
\end{algorithm}
\setlength{\textfloatsep}{0pt}
\section{Experiments}
We evaluate \ours\space using simulated and real-robot scenarios to highlight the importance of high-frequency replanning and demonstrate the effectiveness of our framework. First, we show that our method remains competitive with previous work \cite{chi2023,Ze2024DP3} on the standard Metaworld benchmark. Next, we augment the reach task of MetaWorld with obstacles, which have never been seen during pretraining, and compare our method to an augmented version of DP with guidance. We continue with real-world robot experiments to showcase our method's robustness and generalizability. Finally, we highlight the impact of our design choices by conducting ablation studies.

\subsection{Replanning Frequency Evaluation on MetaWorld}
\begin{table}[h]
\vspace{0.12cm}
\caption{Success rate and replanning frequency (RF) on obstacle free MetaWorld benchmarks for easy (E), intermediate (I), hard (H), and very hard (VH) scenarios. \textbf{Boldface}: best result.}
\label{tb:benchmark_res}
\centering
\resizebox{\columnwidth}{!}{%
\begin{tabular}{c | c | c c c c | c | c} %{\columnwidth}
\toprule
Observation & Method & E (28) & I (11) & H (6) & VH(5)& Avg. & RF (Hz) \\
\midrule
\multirow{2}{*}{State}& DP \cite{chi2023}& $78.9$ & $50.6$ & $68.2$ & \textbf{55.1} & 63.2 & 36.7\\
& \ours     & \textbf{91.0} & \textbf{50.6} & \textbf{75.8} & 47.0 & \textbf{66.1} & $\mathbf{130.9}$\\
% & RTC       & \textbf{93.62} & \textbf{58.11} & \textbf{100} & 73.66 & $\mathbf{148.76 \pm 0.15}$\\
% & DP       & \textbf{81.72} & \textbf{71.11} & \textbf{96.11} & 67.41 & $\mathbf{148.76 \pm 0.15}$\\
\midrule
\multirow{2}{*}{Point Cloud} 
% & DP \cite{chi2023}    & 77.1 & 18.0 & 3.0 & 22.5 & 30.15 & time\\
& DP3\cite{Ze2024DP3}   & 90.9 & 61.6 & 31.7 & 49.0 & 58.3 & 16.1\\
& \ours   & \textbf{93.3} & \textbf{66.9} & \textbf{42.7} & \textbf{71.4} & \textbf{68.6} & $\mathbf{82.3}$\\
\bottomrule
\end{tabular}
}\vspace{0.2cm}
\end{table}

In this section, we evaluate \ours\space on the full suite of MetaWorld tasks \cite{yu2019meta} to confirm its competitiveness to SOTA approaches while achieving higher replanning frequencies. The tasks are categorized into four difficulty levels following \cite{pmlr-v205-seo23a}. As benchmarks are done in obstacle-free environments, all guidance signals are set to zero.

\textbf{Experiments Setup.} 
Expert demonstrations are obtained using the heuristic policies from MetaWorld for both state and point cloud observations. We collect only 10 demonstrations for all tasks following \cite{Ze2024DP3}. To ensure a fair comparison, we maintain the same resolution for images and depth maps as in the previous works. For state-based policies, we benchmark against Diffusion Policy (DP) \cite{chi2023}, and for point cloud observations, against 3D Diffusion Policy (DP3) \cite{Ze2024DP3}. 

We train the state-based models (both DP and \ours) for 3,000 epochs, while the point cloud version of \ours\space is trained for 10,000 epochs. Evaluations are conducted every 200 epochs over 20 episodes, where we report the average top-5 success rates and replanning frequencies. All evaluations are performed on a NVIDIA RTX 4080 SUPER GPU. 

\textbf{Results.} The results are summarized in Table \ref{tb:benchmark_res}. 
For both state-based and point cloud-based models, \ours\space outperforms the SOTA methods with a large margin on success rate across most difficulties. The only exception is for \textit{Very Hard} state-based tasks, which we attribute to insufficient training (3,000 epochs). 

After 10,000 epochs of training for both models, our method achieves a success rate of 76.7\%, outperforming DP, which reaches only 70.9\%. Notably, our replanning frequency is approximately $3.5 \times$ faster than DP and $5 \times$ faster than DP3, highlighting its potential to adapt to rapidly changing environments.

\subsection{Adaptation to Unseen Environmental Changes}
\label{ssec: obs_avoi}

In this section, we consider a common scenario that requires a pre-trained diffusion model to quickly adapt to environmental changes, manipulating objects while avoiding unseen obstacles. Specifically, we consider both static and moving obstacles to test our methods' fast adaptation capability without any re-training or fine-tuning. To the best of our knowledge, this is the first diffusion controller to operate at a replanning frequency that enables continuous interaction with the fast-paced environment. 

\textbf{Guidance Function Design.} We employ the repulsive obstacle function from \cite{song2023loss} as our differentiable training-free guidance loss function. Let $C_{ee}^t$ denote the end-effector's Cartesian coordinates at time step $t$, $C_{obs}^t$ denote the obstacle's center Cartesian coordinates at time step $t$, and $r$ represent the obstacle's radius. For a dynamic obstacle, $C_{obs}^t$ continuously changes. The repulsive loss function is defined as $f = \sum_t \text{sigmoid}\left(-\left(\lVert C_{ee}^t - C_{obs}^t\rVert^2_2 - r\right) \times \lambda\right) \times \omega$, where $\lambda$ and $\omega$ are hyperparameters that control the shape of the function. To generate the guidance signals, we differentiate $f$ with respect to $\hat{A}^0_t$. In our setup, the predicted actions correspond to the end-effector's target position at each step, i.e., $C_{ee}^t$ equals to $\hat{a}_t^0$. Therefore, $f$ is differentiable w.r.t $\hat{A}^0_t$.

\begin{figure}[h]
\vspace{0.2cm}
\centering
%original size 33cm, left, bottom, right, top
%33-12.8 = 20.2cm ==0.94linewdith
%0.94->0.6
\subfloat[][]{
\includegraphics[clip, trim=0cm 0cm 0cm 0cm, width=0.27\linewidth]{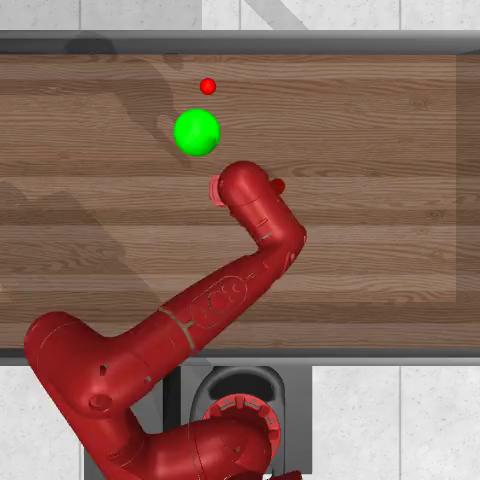}
}
\quad
\subfloat[][]{
\includegraphics[width=0.27\linewidth]{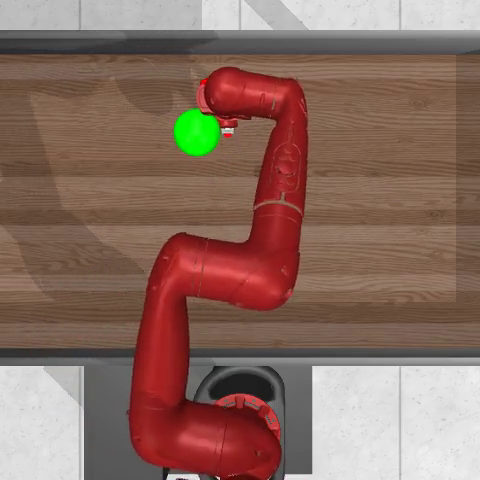}
}
\quad
\subfloat[][]{
\includegraphics[width=0.27\linewidth]{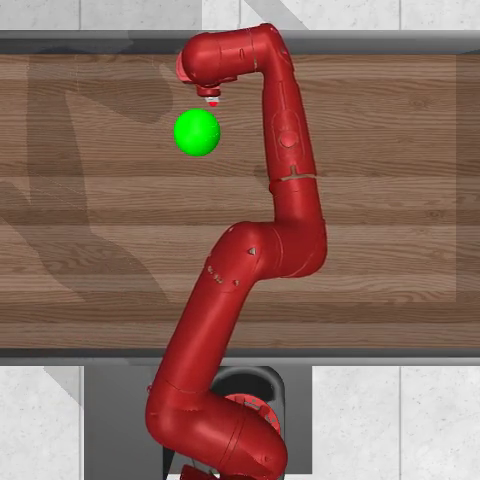}
}\vspace{-0.2cm}
\caption{Modified MetaWorld's \textit{reach-v2} with obstacle avoidance. The end-effector must reach the goal (red dot) without collision with the obstacle (green sphere). Panels (a)–(c) show a \textit{Success} trajectory with a static obstacle.} %
\label{fig: meta_world_figure}
\end{figure}

\textbf{Experiments setup.} 
We collect 4,000 state-based demonstrations in an obstacle-free environment to train all models. In order to benchmark for collision avoidance, we modified MetaWorld's \textit{reach-v2} task. Specifically, a randomly placed spherical obstacle (static or dynamic) simulates an object to be avoided, such as a hand (see Fig. \ref{fig: meta_world_figure} for an example). We augment DP \cite{chi2023} with training-free guidance integration as the baseline model, Guided-DP. Both \ours\space and Guided-DP use the same obstacle avoidance function to generate signals. In our experiments, we set $\lambda=50$, $\omega=100$ and $r = 0.6$.

For obstacle avoidance evaluation, we classify a trajectory as a \textit{Success} if the end-effector reaches the goal without any collisions, and as a \textit{Failure} if it either collides with an obstacle or fails to reach the goal. The models are both trained for 3,000 epochs. We report the average success and failure rates based on 60 trials for static obstacles and 20 trials for dynamic obstacles.

\textbf{Results.} In Table \ref{tb:static_obstacle}, we demonstrate that \ours\space significantly outperforms Guided-DP in both static and dynamic obstacle avoidance. As discussed in Section \ref{ssec:inference}, Guided-DP’s lower replanning frequency leads to worse performance. In static environments, Guided-DP often approaches the obstacle too closely before reacting, resulting in large guidance gradient that causes action prediction divergence. In dynamic environments, it often collides before receiving updated obstacle information.

Although \ours\space performs better overall, its success rate still declines in dynamic environments, primarily because the end-effector deviates too far from the goal during obstacle avoidance, leading to excessively long trajectories that are eventually truncated and resulting in failure. 

\begin{table}[H]
\vspace{-0.2cm}
    \caption{Obstacle avoidance results for static obstacle (60 trials) and dynamic obstacle (20 trials). \textbf{Boldface}: best results.}
    \label{tb:static_obstacle}
    \centering
    \begin{tabular}{c | c c | c c}
    \toprule
        \multirow{2}{*}{Method}  & \multicolumn{2}{c|}{\textit{Static Obstacle}} & \multicolumn{2}{c}{\textit{Dynamic Obstacle}}\\
        & Success$\uparrow$   & Failure $\downarrow$ &Success$\uparrow$   & Failure$\downarrow$\\
        \midrule
        Guided-DP & 15.0   & 85.0  & 10.0 & 90.0 \\
        \midrule
        \ours (ours) & \textbf{63.3}  & \textbf{36.7} & \textbf{45.0} & \textbf{55.0}\\
        \bottomrule
    \end{tabular}
    \vspace{-0.3cm}
\end{table}

\subsection{Real Franka Arm Experiments and Demonstrations}
\label{ssec: real_robot_experiments}
% Given Guided-DP’s unsatisfactory performance in MetaWorld (Table \ref{tb:static_obstacle}), we limit our subsequent real-robot experiments and demonstrations to \ours\space only.

\begin{figure}[h]
\setlength{\belowcaptionskip}{-10pt}
	\includegraphics[width=\linewidth]{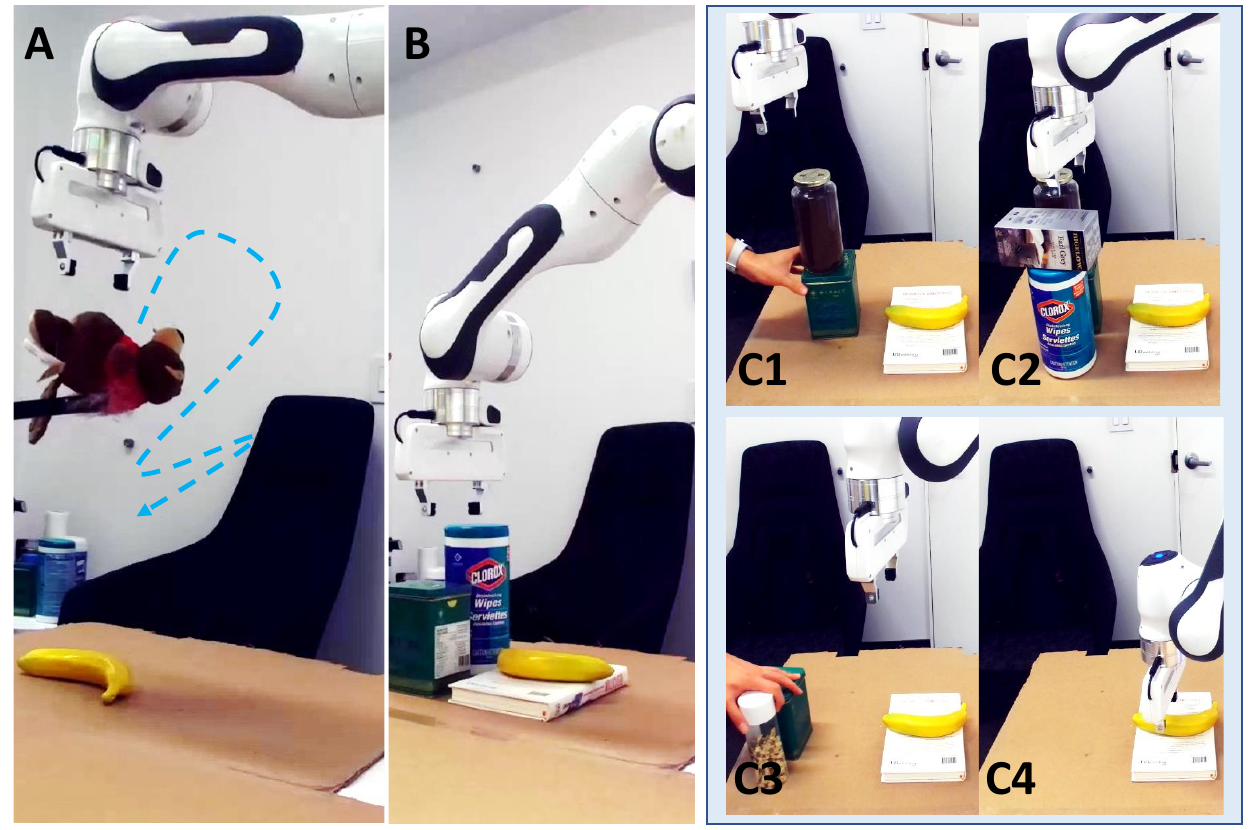}
	\centering
	\caption[real world experiment setup]{Evaluation of \ours\space on real-world manipulation tasks where the robot interacts with obstacles not seen during training. \textbf{A}: Dynamic obstacle avoidance. \textbf{B}: Static obstacle avoidance. \textbf{C1-C4}: Static obstacle avoidance with human interference of stacking and removing different objects.}
	\label{fig:real_exp_steup} \vspace{0.35cm}
\end{figure}

We evaluate \ours\space on real-world grasp-and-lift tasks with obstacle avoidance using a Franka robot. For testing, we introduce various random obstacles, either statically or dynamically, to block the end-effector's path to the object. Two RGBD cameras capture the workspace as point cloud data, which is then transformed into the robot’s base coordinate frame and merged for planning. Off-the-shelf perception modules \cite{liu2023grounding,kirillov2023segany,lin2023sam} are used to segment and locate both the target object and any obstacles within the point cloud. The target object’s geometric coordinates serve as the desired goal position of the end-effector in \ours. We model each obstacle point cloud as a sphere of small radius and apply the obstacle avoidance loss function defined previously to compute the training-free guidance signals for replanning.

To test \ours's training-free replanning frequency, we design three scenarios, as shown in Fig. \ref{fig:real_exp_steup}: (A) dynamic obstacle avoidance, (B) static obstacle avoidance, and (C1-C4) dynamic changes of static obstacles as humans stack or remove objects. In all environments, the task is to grasp and lift a target object, like a banana. We collect 16,000 trajectories with random initial and goal positions, using the ManiSkill2 \cite{gu2023maniskill2} simulator with a Franka arm and heuristic expert controller. Note that all collected data consist solely of trajectories from an initial to a goal position, without any obstacle information. Our dataset exceeds the number of training trajectories used in previous diffusion-based methods because our workspace spans a significantly larger area, demanding broader coverage and leading to an exponential increase in training data requirements \cite{tan2024maniboxenhancingspatialgrasping}. After training, we deploy \ours\space on the real Franka arm for object grasping tasks involving obstacle avoidance.

In the first two environments, \ours\space successfully avoids both static and dynamic obstacles while completing the grasp-and-lift task. The robot consistently maintains a safe distance from the dynamic obstacle, a teddy bear, regardless of its movements. In the third environment, the robot adapts its trajectory effectively despite changes in the scene. For real-world demonstrations of object grasp-and-lift with either static and dynamic obstacle avoidance, please refer to the submitted video demos.

\subsection{Ablation Study}
\label{ssec:ablation}
\begin{figure}[h]
\vspace{-0.2cm}
\centering
%original size 33cm, left, bottom, right, top
%33-12.8 = 20.2cm ==0.94linewdith
%0.94->0.6
\includegraphics[clip, trim=3.2cm 3.9cm 2.05cm 2.95cm, width=0.9\linewidth]{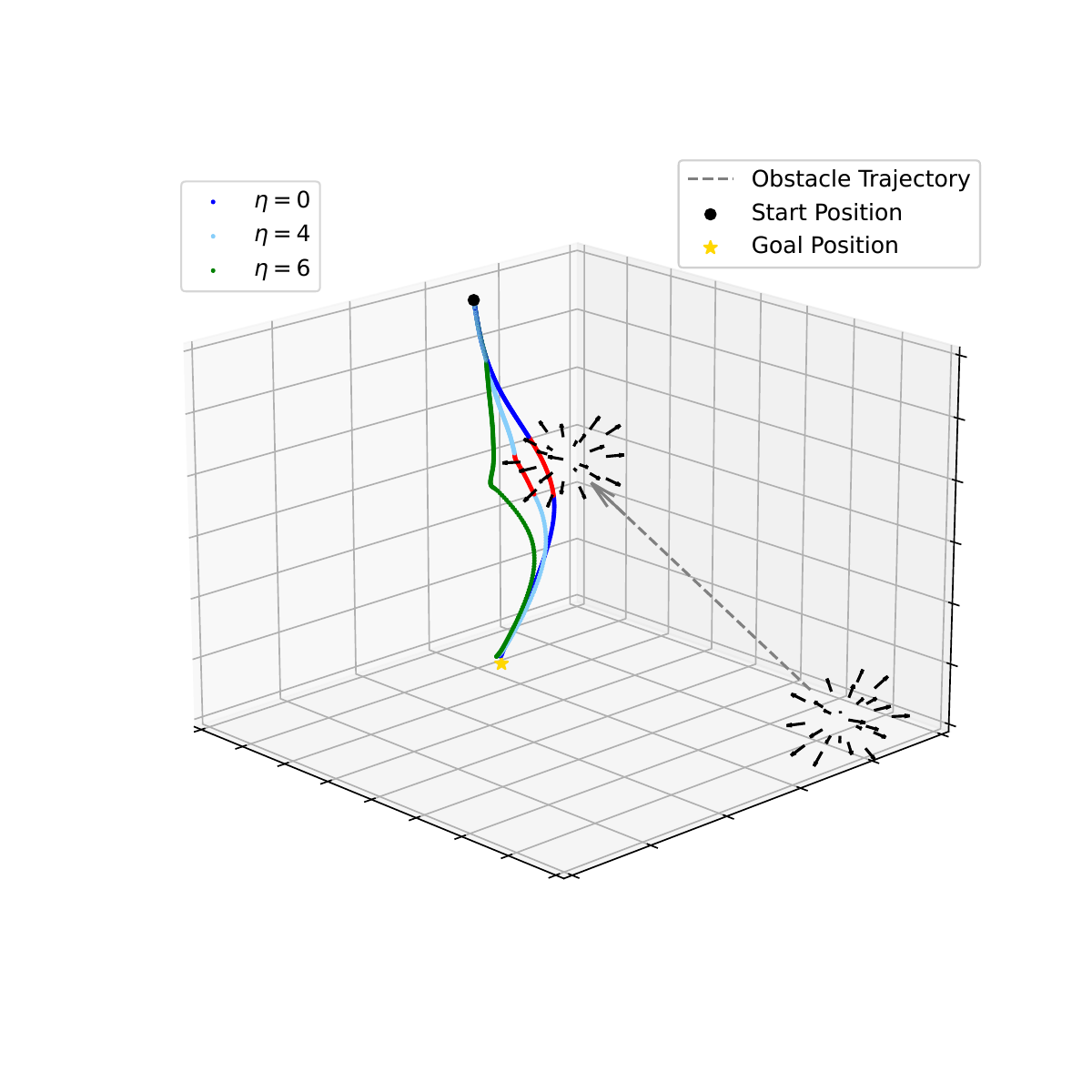}
\vspace{-0.2cm}
\caption{Simulated dynamic obstacle avoidance avoidance trajectories with varying guidance step sizes $\eta$; red segments indicate the collision points with the obstacle.} % from poor action predictions.}
\label{fig:toy_example_a}
\end{figure}

\textbf{Guidance step size.} We employ the trained \ours\space from \ref{ssec: real_robot_experiments} to investigate the influence of guidance step size. This ablation study is conducted using a simulated 3D target-reaching task with dynamic obstacle avoidance.  

Fig. \ref{fig:toy_example_a} illustrates different end-effector trajectories predicted by \ours\space under varying guidance step size $\eta$. Given the same dynamic obstacle moving at constant velocity along a fixed path, larger values of $\eta$ further distancing the end-effector trajectory from the obstacle as expected. In our simulation, the end-effector perfectly avoids an unforeseen dynamic obstacle with a guidance step size of $\eta=6$. We next quantitatively analyze the effect of dynamic obstacle speed in Metaworld with $\eta=6$ .

\begin{figure}[t]
\vspace{0.2cm}
\centering
\includegraphics[clip, trim=0.25cm 0.25cm 0.25cm 0.25cm, width=0.95\linewidth]{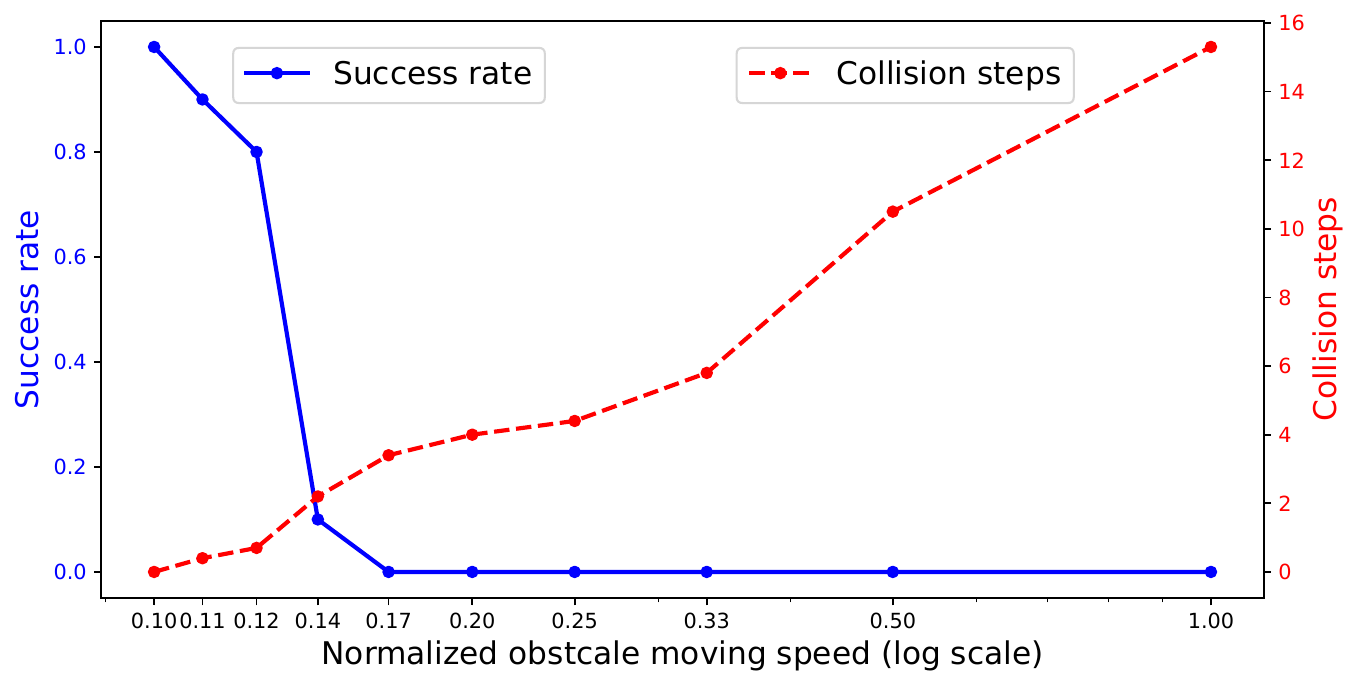}
\caption{Influence of dynamic obstacle moving speed.}
\label{fig:dynamic_speed}\vspace{-0.1cm}
\end{figure}

\textbf{Influence of Dynamic Obstacle Speed.} 
A higher obstacle velocity presents greater challenges for collision avoidance. Here, we test the limits of \ours's replanning ability with various obstacle speeds on the dynamic obstacle avoidance task described in Section \ref{ssec: obs_avoi}. We evaluate the success rate and the number of collision steps at various speeds, as shown in Fig. \ref{fig:dynamic_speed}. We define collision steps as the number of simulation time steps that the end-effector collided with the obstacle. A collision time step of zero  is \textit{Success}. For clarity, the obstacle speed is normalized to a range of 0 to 1. We can observe that as the obstacle speed increases, the success rate declines, and the number of collision steps increases for the trajectories that reach the goal but do not fully avoid the obstacle. When the normalized speed exceeds 0.14, the replanning frequency of \ours\space becomes insufficient, causing the success rate to drop to zero.

\textbf{Mixed Noise Schedule.} The mixed noise schedule controls the balance between monotonic and independent noise injection schedules during training ( see Section \ref{ssec:training}). A ratio of $0.0$ represents a purely monotonic schedule, while a ratio of $1.0$ denotes a purely independent schedule. Adjusting this ratio modulates the probability distribution between the two schemes. 

\begin{table}[H]
    \caption{Results of mixed ratio study. \textbf{Boldface}: best results.}
    \label{tab:ratios}   
    \centering
    \begin{tabular}{l | c c c c c c}
    \toprule
        Ratio  & 0.0 & 0.2 & 0.4 & 0.6 & 0.8 & 1.0 \\
        \midrule
        Success Rate & 21.94 & 56.61 & 58.57 & \textbf{66.31} & 62.57 & 65.44 \\
    \bottomrule
    \end{tabular}\vspace{-0.2cm}
 \end{table}

We evaluate our method on a subset of MetaWorld's easy tasks, using mixed ratios from $0.0$ to $1.0$ in increments of $0.2$ as shown in Table \ref{tab:ratios}. We can see that while an independent noise schedule improves training convergence, the discrepancy between training and inference (which used a monotonic noise schedule) degrades test performance. Thus a mixture of two yields the best results. Empirically, a ratio of $0.6$ yields the best performance.
% We evaluate our method using a subset of MetaWorld's easy tasks with ratios from $0.0$ to $1.0$, in increments of $0.2$. Empirically, we found that a ratio of $0.6$ performs the best.

\section{Conclusion}

In this paper, we proposed a novel diffusion-based framework for training-free high-frequency replanning. Our method integrates a training-free guidance module, allowing generalization to novel dynamic feedbacks during inference. Moreover, our method benefits from an action queuing mechanism, which allows high-frequency replanning. Via empirical evaluation in common and augmented benchmarks as well as real-world scenarios, we showed that the proposed approach achieves state-of-the-art performance while maintaining high replanning frequency, and can effectively handle complex dynamic interactions. Although our study focused on collision avoidance, in theory, the proposed method can incorporate any types of training-free guidance signals. For instance, we can design differentiable functions that enforce a trajectory passage through specific waypoints or dynamically constrain the robot's movement speed.

\bibliographystyle{IEEEtran}
\bibliography{References}
	
\end{document}